\documentclass[paper]{ieice}

\usepackage{graphicx}
\usepackage{amsmath}
\usepackage{amssymb}
\usepackage{amsxtra}
\usepackage{threeparttable}
\usepackage{amsfonts}
\usepackage{mathrsfs}
\usepackage[caption=false, font=footnotesize]{subfig}
\usepackage{arydshln}
\usepackage{cite}
\usepackage{enumitem}
\usepackage{multirow}


\field{A}
\title{A Pseudo Multi-Exposure Fusion Method Using Single Image}
\authorlist{% fill arguments of \authorentry, otherwise error will be caused. 
 \authorentry{Yuma KINOSHITA}{s}{tmu}
 \authorentry{Sayaka SHIOTA}{m}{tmu}
 \authorentry{Hitoshi Kiya}{f}{tmu}
}
\affiliate[tmu]{
  Department of Information and Communication Systems,
  Tokyo Metropolitan University, 191-0065, Tokyo, Japan}
\received{2011}{1}{1}
\revised{2011}{1}{1}
\begin{document}
\setlength{\tabcolsep}{3.0pt}

\maketitle
\begin{summary}
  This paper proposes a novel pseudo multi-exposure image fusion method
  based on a single image.
  Multi-exposure image fusion is used to produce images without saturation regions,
  by using photos with different exposures.
  However, it is difficult to take photos suited for the
  multi-exposure image fusion when we take a photo of dynamic scenes or
  record a video.
  In addition, the multi-exposure image fusion
  cannot be applied to existing images with a single exposure or videos.
  The proposed method enables us to produce pseudo multi-exposure images from a
  single image.
  To produce multi-exposure images, the proposed method
  utilizes the relationship between the exposure values and pixel values,
  which is obtained by assuming that a digital camera has a
  linear response function.
  Moreover, it is shown that the use of a local contrast enhancement method
  allows us to produce pseudo multi-exposure images with higher quality.
  Most of conventional multi-exposure image fusion methods are also applicable to
  the proposed multi-exposure images.
  Experimental results show the effectiveness of the proposed method by
  comparing the proposed one with conventional ones.
\end{summary}

\section{Introduction}
  The low dynamic range (LDR) of the imaging sensors used in modern digital cameras
  is a major factor preventing cameras from capturing images as good as those with human vision.
  For this reason, the interest of high dynamic range (HDR) imaging has
  recently been increasing.
  Various research works on HDR imaging have so far been reported
  \cite{schoberl2013evaluation,chalmers2009high,debevec1997recovering,oh2015robust,
  kinoshita2016remapping,kinoshita2017fast,kinoshita2017fast_trans,huo2016single}.
  The research works are classified into two categories.
  The first one aims to generate HDR images having an extremely wide dynamic range.
  However, HDR display devices are not popular yet due to the high cost of
  the technologies.
  Hence, the second one focuses on tone mapping operations
  which generate standard LDR images from HDR ones
  \cite{murofushi2013integer, murofushi2014integer, dobashi2014fixed}.
  Consequently, in order to generate high quality LDR images via HDR images,
  it is necessary not only to generate HDR ones but also to map them into LDR ones.
  
  To generate LDR images more simply, multi-exposure image fusion methods have been proposed
  \cite{goshtasby2005fusion,mertens2009exposure,saleem2012image,wang2015exposure,
  li2014selectively,sakai2015hybrid,nejati2017fast}.
  The reported fusion methods use a stack of differently exposed images,
  ``multi-exposure images,'' and fuse them to produce an image with high quality.
  The advantage of these methods, compared with the ones via HDR images, is that
  they eliminate three operations:
  generating HDR images, calibrating a camera response function (CRF),
  and preserving the exposure value of each photograph.
  However, the conventional multi-exposure image fusion methods have several problems
  due to the use of a stack of differently exposed images.
  If the scene is dynamic or the camera moves while pictures are
  being captured, the multi-exposure images in the stack will not line up properly
  with one another.
  This misalignment results in ghost-like artifacts in the fused image.
  Although a number of methods have been proposed\cite{li2014selectively,oh2015robust}
  to eliminate these artifacts,
  the effectiveness of these methods is limited because
  it is difficult to apply them to videos.
  In addition, multi-exposure image fusion methods
  cannot be applied to existing images with a single exposure or videos.
  
  Because of such a situation, this paper proposes a novel pseudo multi-exposure image
  fusion method using a single image.
  The proposed method enables us to produce pseudo multi-exposure images from a
  single image and to improve the image quality by fusing them.
  To produce multi-exposure images, the proposed method
  use the relationship between the exposure values and pixel values,
  which is obtained by assuming that a digital camera has a
  linear response function.
  Moreover, the use of a local contrast enhancement method improves the quality of
  the pseudo multi-exposure images.
  Most of conventional multi-exposure image fusion methods are also applicable to
  the proposed pseudo multi-exposure images.
  Furthermore, the proposed method is useful for both reducing
  the number of input images used in conventional fusion ones,
  and improving the quality of multi-exposure images.
  
  We evaluate the effectiveness of the proposed method in terms of the
  quality of generated images by a number of simulations.
  In the simulations, the proposed method is compared with existing
  multi-exposure image fusion methods and typical contrast enhancement methods.
  The results show that the proposed method can produce high quality images,
  as well as conventional fusion methods with multi-exposure images.
  In addition, the proposed method outperforms typical contrast enhancement methods
  in terms of the color distortion.
\section{Preparation}
  Multi-exposure fusion methods use images taken under different exposure conditions,
  i.e., “multi-exposure images.”
  Here we discuss the relationship between exposure values and pixel values.
  For simplicity, we focus on grayscale images.

\subsection{Relationship between exposure values and pixel values}
\begin{figure}[!t]
    \centering
    \includegraphics[width=0.95\linewidth]{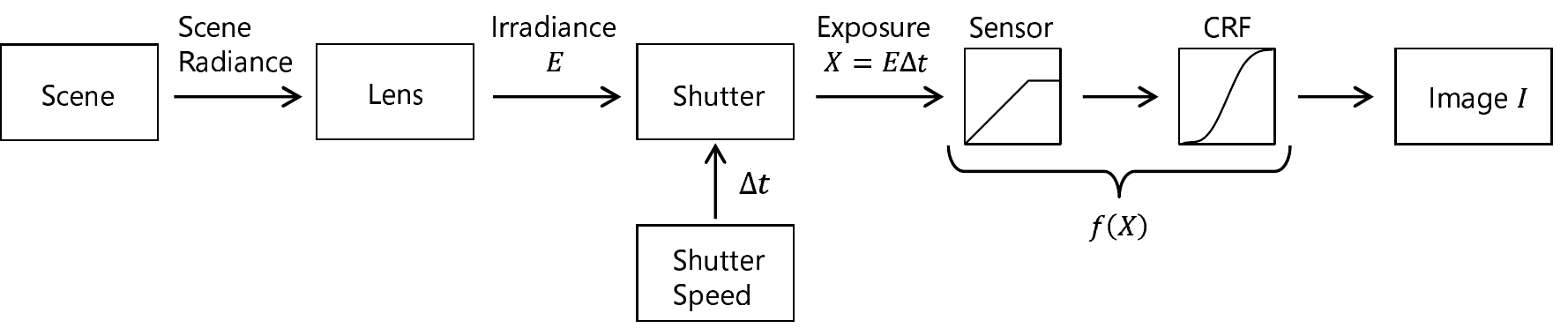}
    \caption{Imaging pipeline of digital camera} \label{fig:camera} 
  \end{figure}

  Figure \ref{fig:camera} shows the imaging pipeline for a digital camera\cite{dufaux2016high}.
  The radiant power density at the sensor, i.e., irradiance $E$,
  is integrated over the time $\Delta t$ the shutter is open, producing an energy density,
  commonly referred to as exposure $X$.
  If the scene is static during this integration,
  exposure $X$ can be written simply as the product of irradiance $E$
  and integration time $\Delta t$ (referred to as "shutter speed"):
  \begin{equation}
    X(p) = E(p)\Delta t ,
    \label{eq:exposure}
  \end{equation}
  where $p=(x,y)$ indicates the pixel at point $(x,y)$.
  A pixel value $I(p) \in [0, 1]$ in the output image $I$ is given by
  \begin{equation}
    I(p) = f(X(p)) ,
    \label{eq:CRF}
  \end{equation}
  where $f$ is a function combining sensor saturation and a camera response function (CRF).
  The CRF represents the processing in each camera which makes
  the final image $I(p)$ look better.

  Camera parameters, such as shutter speed and lens aperture,
  are usually calibrated in terms of exposure value (EV) units,
  and the proper exposure for a scene is automatically selected by the camera.
  The exposure value is commonly controlled by changing the shutter speed
  although it can also be controlled by adjusting various camera parameters.
  Here we assume that the camera parameters except for the shutter speed are fixed.
  Let $0 \mathrm{[EV]}$ and $\Delta t_{0 \mathrm{EV}}$ be the proper exposure value
  and shutter speed under the given conditions, respectively.
  The exposure value $v_i \mathrm{[EV]}$ of an image taken at shutter speed $\Delta t_i$
  is derived from
  \begin{equation}
    v_i = \log_2 \Delta t_i - \log_2 \Delta t_{0 \mathrm{EV}} .
    \label{eq:EV}
  \end{equation}
  From eq. (\ref{eq:exposure}) to eq. (\ref{eq:EV}),
  images $I_{0 \mathrm{EV}}$ and $I_i$ exposed at $0 \mathrm{[EV]}$ and $v_i \mathrm{[EV]}$,
  respectively, are written as
  \begin{align}
    I_{0 \mathrm{EV}}(p) &= f(E(p)\Delta t_{0 \mathrm{EV}}) \label{eq:CRFwithExposure}\\
    I_i(p) &= f(E(p)\Delta t_i) \label{eq:CRFwithExposure2}
    = f(2^{v_i} E(p)\Delta t_{0 \mathrm{EV}}) .
  \end{align}
  Assuming function $f$ is linear,
  we obtain the following relationship between $I_{0 \mathrm{EV}}$ and $I_i$:
  \begin{equation}
    I_i(p) = 2^{v_i} I_{0 \mathrm{EV}}(p) .
    \label{eq:relationship}
  \end{equation}
  Therefore, the exposure can be varied artificially by multiplying $I_{0 \mathrm{EV}}$
  by a constant.
  This ability is used in our proposed pseudo multi-exposure fusion method,
  which is described in the next section.
\section{Proposed pseudo multi-exposure image fusion}
\begin{figure*}[!t]
    \centering
    \includegraphics[clip, width=12cm]{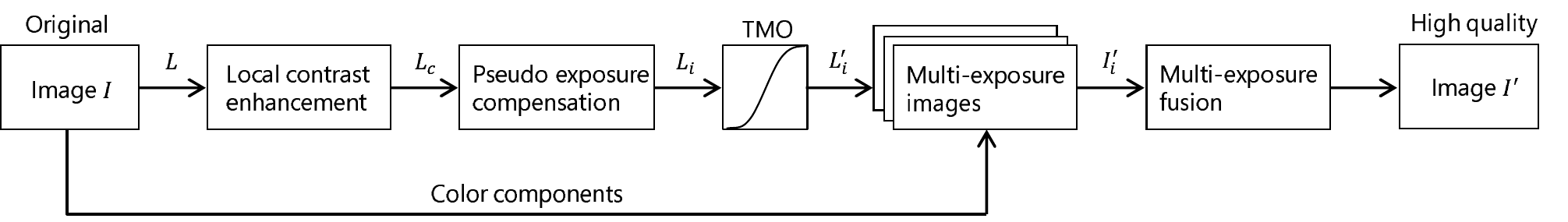}
    \caption{Outline of proposed method \label{fig:PMEF}}
  \end{figure*}

  In this paper, we propose a novel pseudo multi-exposure image fusion method
  which fuses multi-exposure images generated form a single image.
  The outline of the proposed method is shown in Fig. \ref{fig:PMEF}.
  In the proposed method, local contrast enhancement is applied to
  the luminance $L$ calculated from the original image $I$ and then
  pseudo exposure compensation and tone mapping are also applied.
  Next, image $I'$ with improved quality is produced by multi-exposure image fusion.
\subsection{Local contrast enhancement}
  If pseudo multi-exposure images are generated form a single image,
  the quality of an image fused from them will be lower than
  that of an image fused from genuine multi-exposure images.
  Therefore, the dodging and burning algorithm is used to enhance
  the local contrast \cite{huo2013dodging}.
  The algorithm is given by
  \begin{equation}
    L_c(p) = \frac{L^2(p)}{L_a(p)},
    \label{eq:dodgingAndBurning}
  \end{equation}
  where $L_a(p)$ is the local average of luminance $L(p)$ around pixel $p$.
  It is obtained by applying a low-pass filter to $L(p)$.
  Here, a bilateral filter is used for this purpose.

  $L_a(p)$ is calculated using the bilateral filter
  \begin{equation}
    L_a(p) = \frac{1}{c(p)}
              \sum_{q \in \Omega}
                L(q) g_{\sigma_1}(q-p) g_{\sigma_2}(L(q) - L(p)),
    \label{eq:bilateral}
  \end{equation}
  where $\Omega$ is the set of all pixels, and $c(p)$ is a normalization term such as
  \begin{equation}
    c(p) = \sum_{q \in \Omega} g_{\sigma_1}(q-p) g_{\sigma_2}(L(q) - L(p)),
    \label{eq:normalizingConst}
  \end{equation}
  where $g_{\sigma}$ is a Gaussian function given by
  \begin{equation}
    g_{\sigma}(p | p=(x,y)) = C_{\sigma}\exp \left( -\frac{x^2 + y^2}{\sigma^2} \right)
    \label{eq:gaussian}
  \end{equation}
  using a normalization factor $C_{\sigma}$.
  Parameters $\sigma_1 = 16, \sigma_2 = 3/255$ are set in accordance with \cite{huo2013dodging}.
\subsection{Pseudo exposure compensation}
  The pseudo exposure compensation consists of two steps:
  estimating luminance $L_{0 \mathrm{EV}}$ from $L_c$
  and calculating luminance $L_i (1 \le i \le N, i \in \mathbb{N})$ of the $i$th image,
  where $L_{0 \mathrm{EV}}$ is the luminance of the properly exposed image
  i.e. with $0 \mathrm{[EV]}$, and $N$ is the number of pseudo multi-exposure images
  produced by the proposed method.

  In the first step, there are two approaches A and B
  to estimate the luminance $L_{0 \mathrm{EV}}$.
  Approach A estimates $L_{0 \mathrm{EV}}$ on the basis of automatic exposure algorithms
  in digital cameras,
  so that it enables us to avoid color distortions between
  a resulting image and the original image.
  On the other hand, approach B estimates $L_{0 \mathrm{EV}}$
  by using all luminance values of the scene
  unlike the automatic exposure algorithms
  which generally use luminance values in specific area of the scene.
  Hence, approach B allows us to strongly enhance the contrast in all image regions.
  Note that approach A is only available
  when the exposure value $v \mathrm{[EV]}$ of the original image $I$ is known.
  In contrast, approach B is available regardless
  whether the exposure value $v \mathrm{[EV]}$ of $I$ is known or not.
  \begin{description}[style=nextline,font=\mdseries,leftmargin=0pt,listparindent=2em,parsep=1pt]
    \item[A. Estimating $L_{0 \mathrm{EV}}$ with exposure value $v$]
      In approach A, according to eq. (\ref{eq:relationship}),
      $L_{0 \mathrm{EV}}$ is estimated as
      \begin{equation}
        L_{0 \mathrm{EV}}(p) = 2^{-v} L_c(p).
        \label{eq:knownEV}
      \end{equation}
    \item[B. Estimating $L_{0\mathrm{EV}}$ without exposure value $v$]
      In approach B, we map the geometric mean $\overline{L}_c$ of luminance $L_c$ to
      middle-gray of the displayed image, or 0.18 on a scale from zero to one,
      as in \cite{reinhard2002photographic},
      where the geometric mean of the luminance values indicates
      the approximate brightness of the image.
      
      The luminance $L_{0\mathrm{EV}}$ is derived from
      \begin{equation}
        L_{0\mathrm{EV}}(p) = \frac{0.18}{\overline{L}_c} L_c(p)
        \label{eq:unknownEV}
      \end{equation}
      where the geometric mean $\overline{L}_c$ of $L_c(p)$ is calculated using
      \begin{equation}
        \overline{L}_c = \exp{\left(\frac{1}{|\Omega|} \sum_{p \in \Omega}
          \log{L_c(p)}\right)}.
        \label{eq:geoMean}
      \end{equation}
      If eq. (\ref{eq:geoMean}) has singularities at some pixels i.e. $L_c(p)=0$, 
      $\overline{L}_c$ is calculated by
      \begin{equation}
        \overline{L}_c =
          \exp{
            \left(\frac{1}{|\Omega|}
              \left(
                \sum_{p \notin B} \log{L_c(p)}
                + \sum_{p \in B} \log{\epsilon}
              \right)
            \right)
          }
        \label{eq:geoMeanEps}
      \end{equation}
      where $B = \{p | L_c(p)=0\}$ and $\epsilon$ is a small value.
  \end{description}

  The second step of the pseudo exposure compensation is carried out
  according to eq. (\ref{eq:relationship}).
  The luminance $L_i$ of the $i$th image $I_i$ is obtained by
  \begin{equation}
    L_i(p) = 2^{v_i} L_{0\mathrm{EV}}(p),
    \label{eq:constMultiplication}
  \end{equation}
  so that the image $I_i$ could have the exposure value $v_i \mathrm{[EV]}$.
  To generate high quality images,
  multi-exposure images should represent bright, middle and dark
  regions of the original image $I$, respectively.
  Since the image having $0 \mathrm{[EV]}$ represents the middle region clearly, 
  a negative value, zero and a positive value should be used as the parameters $v_i$.
  In this paper, we use $N = 3$, and $v_i = -1, 0, +1 \mathrm{[EV]}$.
\subsection{Tone mapping}
  Since the luminance value $L_i(p)$ calculated by the pseudo exposure compensation
  often exceeds the maximum value of the common image format.
  Pixel values might be lost due to truncation of the values.
  This problem is overcome, by using a tone mapping operation
  to fit the luminance value into the interval $[0, 1]$.

  The luminance $L'_i$ of a pseudo multi-exposure image is obtained,
  by applying a tone mapping operator $F_i$ to $L_i$:
  \begin{equation}
    L'_i(p) = F_i(L_i(p)).
    \label{eq:TM}
  \end{equation}
  Reinhard's global operator is used here as tone mapping operator $F_i$
  \cite{reinhard2002photographic}.
  
  Reinhard's global operator is given by
  \begin{equation}
    F_i(L(p)) = \frac{L(p)\left(1 + \frac{L(p)}{L^2_{white_i}} \right)}{1 + L(p)},
    \label{eq:reinhardTMO}
  \end{equation}
  where parameter $L_{white_i} > 0$ determines luminance value $L(p)$
  as $L'(p) = F_i(L(p)) = 1$.
  Note that Reinhard's global operator $F_i$ is a monotonically increasing function.
  Here, let $L_{white_i} = \max L_i(p)$.
  We obtain $L'_i(p) \le 1$ for all $p$.
  Therefore, truncation of the luminance values can be prevented.

  Combining $L'_i$,
  luminance $L$ of the original image $I$,
  and RGB pixel values $C(p) \in \{R(p), G(p), B(p)\}$ of $I$,
  we obtain RGB pixel values $C'_i(p) \in \{R'_i(p), G'_i(p), B'_i(p)\}$ of
  pseudo multi-exposure images $I'_i$:
  \begin{equation}
    C'_i(p) = \frac{L'_i(p)}{L(p)}C(p).
    \label{eq:color}
  \end{equation}
\subsection{Fusion of pseudo multi-exposure images}
  Pseudo multi-exposure images $I'_i$ can be used as input
  for any multi-exposure image fusion method.
  While numerous methods for fusing images have been proposed,
  here we use those of Mertens et al. \cite{mertens2009exposure},
  Sakai et al. \cite{sakai2015hybrid}, and Nejati et al. \cite{nejati2017fast}.
  A final image $I'$ is produced using
  \begin{equation}
    I' = \mathscr{F}(I'_1, I'_2, \cdots, I'_N),
    \label{eq:fusion}
  \end{equation}
  where $\mathscr{F}(I_1, I_2, \cdots, I_N)$ indicates a function to fuse $N$ images
  $I_1, I_2, \cdots, I_N$ into a single image.
\subsection{Proposed procedure}
  The procedure for generating an image $I'$ from the original image $I$
  by the proposed method is summarized as follows (see Fig. \ref{fig:PMEF}).
  \begin{enumerate}[nosep]
    \item Calculate luminance $L$ of the original image $I$.
    \item Calculate $L_c$ by using eq. (\ref{eq:dodgingAndBurning}) to eq. (\ref{eq:gaussian}).
    \item Calculate $L_i$ according to eq. (\ref{eq:constMultiplication}).
      \begin{enumerate}[label=Approach \Alph*.,leftmargin=*]
        \item Calculate $L_{0\mathrm{EV}}$ by eq. (\ref{eq:knownEV}).
        \item Calculate $L_{0\mathrm{EV}}$ by
          eqs. (\ref{eq:unknownEV}) and (\ref{eq:geoMeanEps}).
      \end{enumerate}
    \item Calculate luminance values $L'_i$ of pseudo multi-exposure images $I'_i$
      from eqs. (\ref{eq:TM}) and (\ref{eq:reinhardTMO}).
    \item Generate $I'_i$ according to eq. (\ref{eq:color}).
    \item Obtain an image $I'$ with a multi-exposure image fusion method $\mathscr{F}$
      as in eq. (\ref{eq:fusion}).
  \end{enumerate}
\section{Simulation}
  Using two simulations, ``Simulation 1'' and ``Simulation 2,''
  we evaluated the quality of the images produced by the proposed method,
  the three fusion methods mentioned above,
  and typical single image based contrast enhancement methods, i.e.
  the histogram equalization (HE),
  the contrast limited adaptive histograph equalization (CLAHE) \cite{zuiderveld1994contrast},
  and the contrast-accumulated histogram equalization (CACHE) \cite{wu2017contrast}.
\subsection{Comparison with conventional methods}
  To evaluate the quality of the images produced by each method,
  objective metrics are needed.
  Typical metrics such as the peak signal to noise ratio (PSNR)
  and the structural similarity index (SSIM) are not suitable for this purpose
  because they use the target image with the highest quality as a reference one.
  We therefore used TMQI\cite{yeganeh2013objective} and
  CIEDE2000\cite{sharma2005ciede2000} as the metrics as they do not require
  any reference images.

  TMQI represents the quality of images tone mapped from an HDR image;
  the index incorporates structural fidelity and statistical naturalness.
  An HDR image is used as a reference to calculate structural fidelity.
  Any references are not needed to calculate statistical naturalness.
  Since the processes of tone mapping and photographing are similar,
  TMQI is also useful for evaluating photographs.
  CIEDE2000 represents
  the distance in a color space between two images.
  We used CIEDE2000 to evaluate the color distortion caused by the proposed method.
\subsection{Simulation conditions}
  \subsubsection{Simulation 1 (using HDR images)}
\begin{figure}[!t]
  \centering
  \subfloat[$I_{M1}$ \newline ($v_{M1}=-1\mathrm{[EV]}$)]{
    \includegraphics[width=2.6cm]{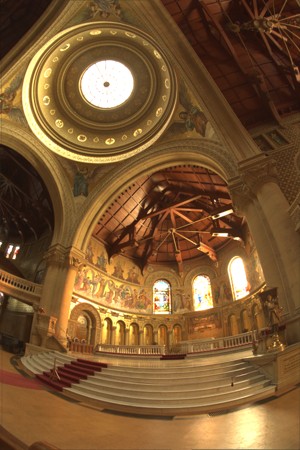}
    \label{fig:OrgM1EV}}
  \subfloat[$I_{M2}$ \newline ($v_{M2}=0\mathrm{[EV]}$)]{
    \includegraphics[width=2.6cm]{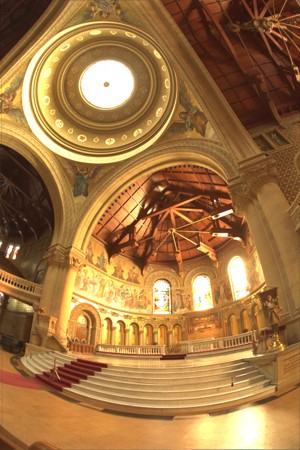}
    \label{fig:Org0EV}}
  \subfloat[$I_{M3}$ \newline ($v_{M3}=+1\mathrm{[EV]}$)]{
    \includegraphics[width=2.6cm]{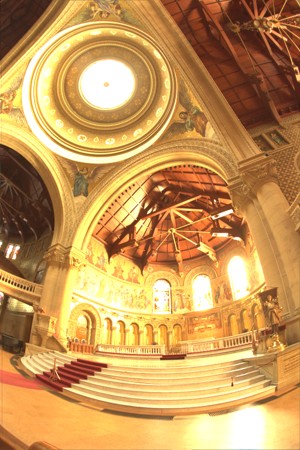}
    \label{fig:OrgP1EV}}\\
  \caption{Examples of multi-exposure images $I_{Mk}$ (Memorial) mapped from $I_H$}
  \label{fig:orgImages}
\end{figure}

  In Simulation 1, HDR images were used to prepare the input images for the proposed method.
  The following procedure was carried out to evaluate the effectiveness of the proposed method.
  \begin{enumerate}[nosep]
    \item Map HDR image $I_H$ to three multi-exposure images $I_{Mk}, k = 1,2,3$
      with exposure values $v_{Mk} = k-2\mathrm{[EV]}$
      by using a tone mapping operator (see Fig. \ref{fig:orgImages}).
    \item Obtain $I'$ from $I$ according to the proposed procedure as in 3.5,
      under $I=I_{M2}$ having $v_{M2} = 0\mathrm{[EV]}$.
    \item Compute TMQI values between $I'$ and $I_H$.
    \item Compute CIEDE2000 values as an error measure between $I'$ and $I_{M2}$.
  \end{enumerate}
  In step 1), the tone mapping operator corresponds to function $f$
  in eqs. (\ref{eq:CRFwithExposure})
  and (\ref{eq:CRFwithExposure2}) (see Fig. \ref{fig:camera}).
  As assumed for eq. (\ref{eq:relationship}),
  a linear operator was used as the tone mapping operator.
  In addition, the properly exposed image, having $0\mathrm{[EV]}$,
  for each scene was defined as an image in which the geometric mean of the luminance
  equals to 0.18.

  We used 60 HDR images selected from available online databases
  \cite{openexrimage,anyherehdrimage}.
  \subsubsection{Simulation 2 (photographing directly)}
\begin{figure}[!t]
  \centering
  \subfloat[$I_{M1}$ \newline ($v_{M1}=-1.3\mathrm{[EV]}$)]{
    \includegraphics[width=2.6cm]{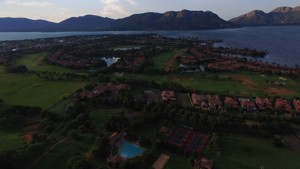}
    \label{fig:estateM1EV}}
  \subfloat[$I_{M2}$ \newline ($v_{M2}=0\mathrm{[EV]}$)]{
    \includegraphics[width=2.6cm]{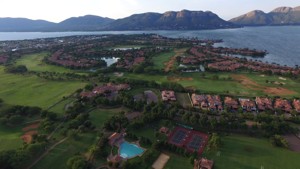}
    \label{fig:estate0EV}}
  \subfloat[$I_{M3}$ \newline ($v_{M3}=+1.3\mathrm{[EV]}$)]{
    \includegraphics[width=2.6cm]{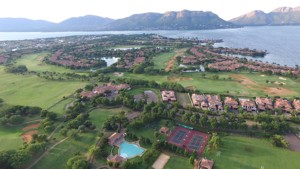}
    \label{fig:estateP1EV}}\\
  \caption{Examples of multi-exposure images $I_{Mk}$ (Estate rsa) for Simulation 2}
  \label{fig:estate}
\end{figure}

  In Simulation 2, four photographs taken by Canon EOS 5D Mark II camera
  and eight photographs selected from an available online database \cite{easyhdr}
  were directly used as input images $I_{Mk}$
  (see Fig. \ref{fig:estate}).
  Since there were no HDR images for Simulation 2,
  the first step in Simulation 1 was not needed.
  In addition, structural fidelity in TMQI could not be calculated
  due to the non-use of HDR images.
  Thus, we used only statistical naturalness in TMQI as a metric.
\subsection{Simulation results}
  Here, the effectiveness of the proposed method is discussed
  on the basis of objective assessments.
\subsubsection{Simulation 1}
\begin{figure}[!t]
  \centering
  \subfloat[Mertens\cite{mertens2009exposure}]{
    \includegraphics[width=2.6cm]{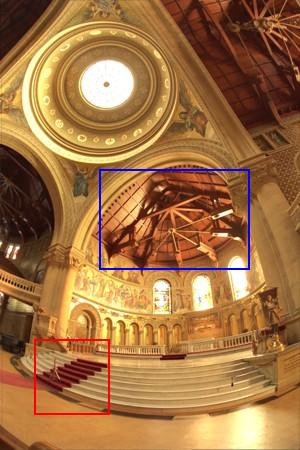}
    \label{fig:Mertens}}
  \subfloat[Sakai\cite{sakai2015hybrid}]{
    \includegraphics[width=2.6cm]{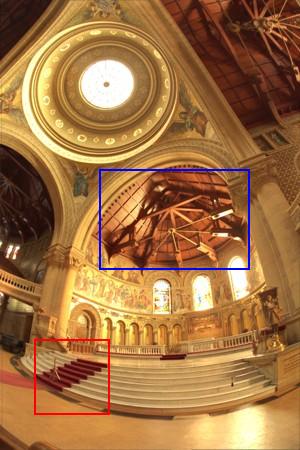}
    \label{fig:Yoshida}}
  \subfloat[Nejati\cite{nejati2017fast}]{
    \includegraphics[width=2.6cm]{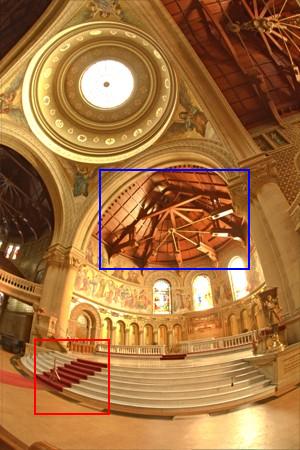}
    \label{fig:Nejati}}\\
  \subfloat[HE]{
    \includegraphics[width=2.6cm]{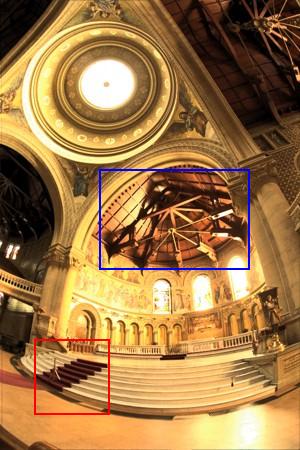}
    \label{fig:HE}}
  \subfloat[CLAHE\cite{zuiderveld1994contrast}]{
    \includegraphics[width=2.6cm]{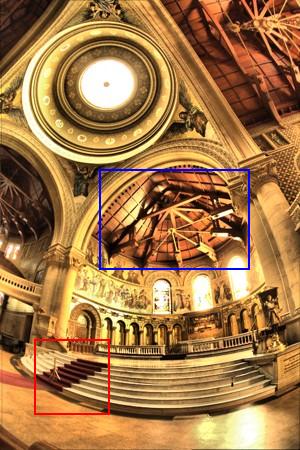}
    \label{fig:CLAHE}}
  \subfloat[CACHE\cite{wu2017contrast}]{
    \includegraphics[width=2.6cm]{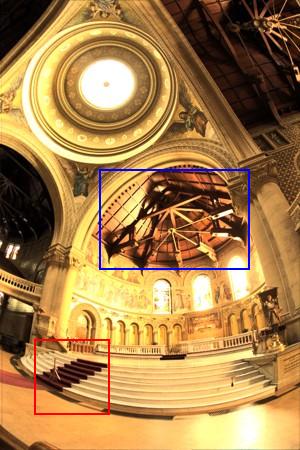}
    \label{fig:CACHE}}\\
  \subfloat[Proposed (B) \newline with \cite{mertens2009exposure}]{
    \includegraphics[width=2.6cm]{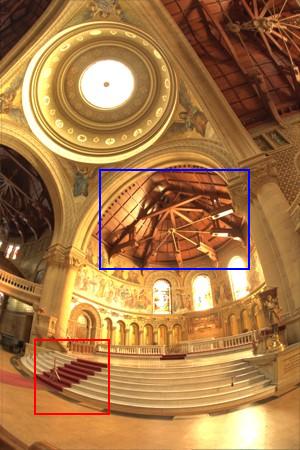}
    \label{fig:PMEFMertens}}
  \subfloat[Proposed (B) \newline with \cite{sakai2015hybrid}]{
    \includegraphics[width=2.6cm]{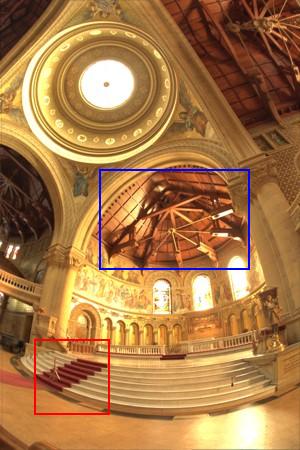}
    \label{fig:PMEFSakai}}
  \subfloat[Proposed (B) \newline with \cite{nejati2017fast}]{
    \includegraphics[width=2.6cm]{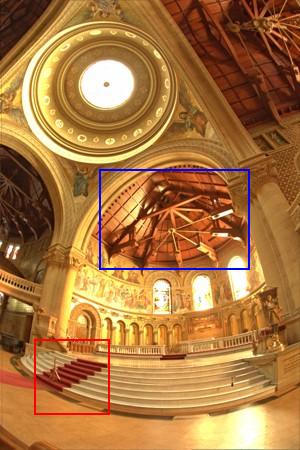}
    \label{fig:PMEFNejati}}\\
  \caption{Images $I'$ generated from image ``Memorial''}
  \label{fig:results}
\end{figure}

  Tables \ref{tab:HDRTMQI}, \ref{tab:HDRNaturalness} and \ref{tab:HDRCIEDE}
  summarize TMQI score, statistical naturalness score, and CIEDE2000 score
  for Simulation 1, respectively.
  For TMQI $\in [0, 1]$ (and statistical naturalness $\in [0, 1]$),
  a larger value means higher quality.
  For CIEDE2000 $\in [0, \infty)$,
  a smaller value intends that the color difference between two images is smaller.
  \begin{description}[style=nextline,font=\mdseries,leftmargin=0pt,listparindent=2em,parsep=1pt]
    \item[a) Comparison with multi-exposure fusion methods]
      Table \ref{tab:HDRTMQI} shows the results of evaluating
      three multi-exposure fusion methods (MEF),
      three conventional contrast enhancement methods (CE),
      and the proposed method, in terms of TMQI,
      where the proposed method has six variations.
      Here CE and the proposed method utilized a single image $I_{M2}$ having $0 \mathrm{[EV]}$
      as the input image, although MEF used three multi-exposure images
      $I_{M1}, I_{M2}$ and $I_{M3}$ as input ones.
      By comparing MEF with approach A and B
      (e.g. comparing MEF \cite{mertens2009exposure}
      with the proposed method using \cite{mertens2009exposure}),
      it is confirmed that both approach A and B provide higher TMQI scores than MEF,
      even though the proposed ones used a single image as an input image.
      Statistical naturalness scores (in Table \ref{tab:HDRNaturalness})
      also show a similar trend to Table \ref{tab:HDRTMQI}.

      By considering CIEDE2000 scores in Table \ref{tab:HDRCIEDE},
      it is also confirmed that approach A has better CIEDE scores
      than MEF.

      Figure \ref{fig:results} shows an example of images generated by each method.
      In this figure, the results of approach A are not shown because
      there were few visual differences between approach A and approach B.
      This is because exposure values of input images were determined in the same way as
      that utilized in approach B for estimating $L_{0 \mathrm{EV}}$
      (given by eq. \ref{eq:unknownEV}), in Simulation 1.
      From the figure, it is confirmed that
      the proposed method can produce an image with almost the same as ones
      fused by MEF.

      These results demonstrate that the proposed method is effective as well as MEF.
      Moreover, CIEDE2000 scores denote that approach A
      can produce images with higher quality, in terms of the color distortion,
      than approach B.
    \item[b) Comparison with contrast enhancement methods]
      Contrast enhancement also allows us to enhance the quality of images from a single image.
      To clearly show the effectiveness of the proposed method,
      we compared the proposed method with typical contrast enhancement methods.
      
      Contrast enhancement methods provided higher TMQI and statistical naturalness scores
      than that of the proposed ones as shown in
      Tables \ref{tab:HDRTMQI} and \ref{tab:HDRNaturalness}.
      Especially, CACHE which is the state-of-the-art method has the best scores in all methods.
      However, they have the worst CIEDE2000 scores (see Table \ref{tab:HDRCIEDE}).
      The result means that the use of a contrast enhancement method would
      produce some serious color distortion.
      By comparing Fig. \ref{fig:results} with Fig. \ref{fig:orgImages},
      it is also confirmed that contrast enhancement methods bring color distortion,
      e.g. the carpet on stairs (boxed by red line).
      In addition, since contrast enhancement methods aim to maximize image contrast,
      the resulting images sometimes have unnatural contrast
      due to over-enhancement
      (see regions boxed by blue line in Fig. \ref{fig:results}).
      By contrast, the proposed method can prevent both the color distortion
      and the over-enhancement.
      Therefore, the proposed methods outperforms contrast enhancement methods
      in terms of the color distortion
      and the over-enhancement.

      The results of Simulation 1 show that the proposed method
      enables us to produce high-quality images as well as conventional MEF,
      even when a single image is used as an input image.
      Besides, the proposed method also outperforms CE
      in therms of the color distortion and the over-enhancement.
      Comparison between approach A and B demonstrate that approach A can provide better
      CIEDE2000 scores than approach B,
      although approach B can strongly enhance the contrast of images as described later.
  \end{description}
\subsubsection{Simulation 2}
\begin{figure}[!t]
  \centering
  \subfloat{
    \includegraphics[width=.4\columnwidth]{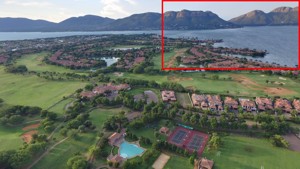}
  }
  \subfloat{
    \includegraphics[width=.4\columnwidth]{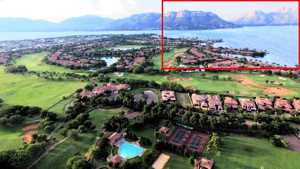}
  }\\
  \vspace{-2mm}
  \addtocounter{subfigure}{-2}
  \subfloat[Mertens\cite{mertens2009exposure}]{
    \includegraphics[width=.4\columnwidth]{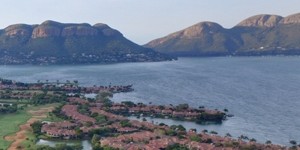}
    \label{fig:MertensCorridor1}}
  \subfloat[HE]{
    \includegraphics[width=.4\columnwidth]{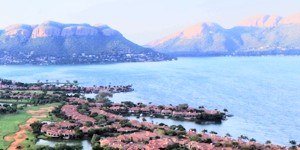}
    \label{fig:HECorridor1}}\\
  \subfloat{
    \includegraphics[width=.4\columnwidth]{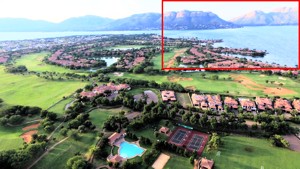}
  }
  \subfloat{
    \includegraphics[width=.4\columnwidth]{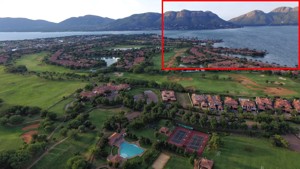}
  }\\
  \vspace{-2mm}
  \addtocounter{subfigure}{-2}
  \subfloat[CACHE\cite{wu2017contrast}]{
    \includegraphics[width=.4\columnwidth]{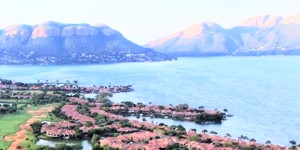}
    \label{fig:CACHECorridor1}}
  \subfloat[Proposed (A) \newline with \cite{mertens2009exposure}]{
    \includegraphics[width=.4\columnwidth]{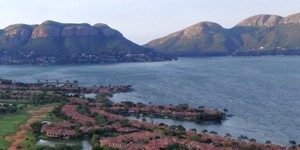}
    \label{fig:PMEFMertensCorridor1}}\\
  \subfloat{
    \includegraphics[width=.4\columnwidth]{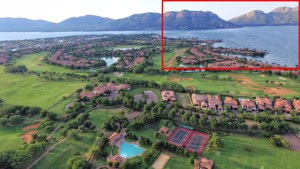}
  }\\
  \addtocounter{subfigure}{-1}
  \subfloat[Proposed (B) \newline with \cite{mertens2009exposure}]{
    \includegraphics[width=.4\columnwidth]{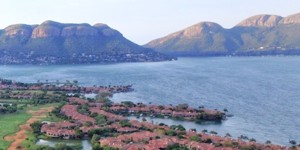}
    \label{fig:PMEFMertensCorridor1}}\\
  \caption{Images $I'$ generated from image ``Estate rsa'' (top)
    and zoom-in views of their upper right corner (bottom).}
  \label{fig:resultsCamera}
\end{figure}

  In Simulation 2, statistical naturalness scores also show a similar trend to Simulation 1
  (see Table \ref{tab:CameraNaturalness}).
  Besides, Table \ref{tab:CameraCIEDE} shows that
  proposed methods using approach B as in 3.2
  has worse CIEDE2000 scores than CLAHE and CACHE.
  This is due to the difference of estimating method
  for $L_{0\mathrm{EV}}$ between digital cameras and the proposed method
  using approach B.
  In approach B, estimated $L_{0\mathrm{EV}}$ is differ from ones estimated by digital cameras.
  As a result, brightness of images produced by the proposed one
  using approach B is differ substantially
  from the input image as shown in Fig. \ref{fig:resultsCamera}.
  On the other hand, approach A enables us to avoid color distortions since
  it estimates $L_{0\mathrm{EV}}$ by using exposure values calculated by digital cameras.
  Thus, approach A has the lowest CIEDE2000 scores in the methods
  (see Table \ref{tab:CameraCIEDE}).
  
  From Fig. \ref{fig:resultsCamera},
  it is also confirmed that CE methods
  (He and CACHE) cause
  the loss of details in bright regions boxed by red line.
  This is due to the fact that these CE methods decrease
  the number of gradations assigned for bright regions, to enhance dark regions.
  By contrast, both approaches A and B
  can enhance images without the loss of details,
  as well as conventional MEF.

  From these results, the proposed method enables us to generate images with high quality,
  as well as conventional MEF, from a single image.
  In addition, approach A outperforms
  typical contrast enhancement methods in terms of the color distortion.
  On the other hand, approach B can strongly enhance
  the contrast of images without loss of details,
  unlike conventional CE methods.
\begin{table*}[!t]
  \centering
  \caption{Experimental results for Simulation 1 (TMQI).
  ``MEF,'' and ``CE'' indicate
  multi-exposure fusion and contrast enhancement, respectively.}
  {\footnotesize
  \begin{tabular}{l|l|lll|lll|ll|ll|ll} \hline \hline
    \multirow{3}{*}{Methods} & \multirow{3}{10mm}{Input image} & \multicolumn{3}{c|}{MEF} &
    \multicolumn{3}{c|}{CE} & \multicolumn{6}{c}{Proposed}\\\cline{3-14}
    & & \multicolumn{1}{c}{\cite{mertens2009exposure}} &
    \multicolumn{1}{c}{\cite{sakai2015hybrid}} & \multicolumn{1}{c|}{\cite{nejati2017fast}} &
    \multicolumn{1}{c}{HE} & \multicolumn{1}{c}{\cite{zuiderveld1994contrast}} &
    \multicolumn{1}{c|}{\cite{wu2017contrast}} &
    \multicolumn{2}{c|}{\cite{mertens2009exposure}} &
    \multicolumn{2}{c|}{\cite{sakai2015hybrid}} &
    \multicolumn{2}{c}{\cite{nejati2017fast}}\\
    &&&&&&&& \multicolumn{1}{c}{A} & \multicolumn{1}{c|}{B} &
    \multicolumn{1}{c}{A} & \multicolumn{1}{c|}{B} & \multicolumn{1}{c}{A} &
    \multicolumn{1}{c}{B}\\
    \hdashline
AtriumNight & 0.8388 & 0.8514 & 0.8510 & 0.8402 & 0.8536 & 0.8236 & \textbf{0.8710} & 0.8579 & 0.8604 & 0.8576 & 0.8601 & 0.8449 & 0.8473 \\
MtTamWest & 0.7189 & 0.7784 & 0.7785 & 0.7718 & 0.7838 & \textbf{0.8838} & 0.8133 & 0.7990 & 0.8215 & 0.7964 & 0.8182 & 0.7885 & 0.8139 \\
SpheronNapa & 0.7239 & 0.7485 & 0.7483 & 0.7515 & 0.7423 & \textbf{0.7933} & 0.7734 & 0.7633 & 0.7670 & 0.7624 & 0.7660 & 0.7572 & 0.7610 \\
Memorial & 0.8404 & 0.8427 & 0.8429 & 0.8396 & 0.8381 & 0.7872 & 0.8415 & 0.8461 & 0.8522 & 0.8473 & \textbf{0.8538} & 0.8379 & 0.8438 \\
Rend 11 & 0.7932 & 0.8242 & 0.8231 & 0.8142 & 0.8649 & 0.8908 & \textbf{0.8994} & 0.8312 & 0.8563 & 0.8303 & 0.8552 & 0.8207 & 0.8474 \\\hdashline
Average & \multirow{2}{*}{0.7830} & \multirow{2}{*}{0.8090} & \multirow{2}{*}{0.8088} & \multirow{2}{*}{0.8034} & \multirow{2}{*}{0.8165} & \multirow{2}{*}{0.8358} & \multirow{2}{*}{\textbf{0.8397}} & \multirow{2}{*}{0.8195} & \multirow{2}{*}{0.8315} & \multirow{2}{*}{0.8188} & \multirow{2}{*}{0.8307} & \multirow{2}{*}{0.8099} & \multirow{2}{*}{0.8227} \\
(5 images) &  &  &  &  &  &  &  &  &  &  &  &  &  \\\hdashline
Average & \multirow{2}{*}{0.8088} & \multirow{2}{*}{0.8151} & \multirow{2}{*}{0.8151} & \multirow{2}{*}{0.8130} & \multirow{2}{*}{0.8376} & \multirow{2}{*}{0.8248} & \multirow{2}{*}{\textbf{0.8581}} & \multirow{2}{*}{0.8294} & \multirow{2}{*}{0.8355} & \multirow{2}{*}{0.8290} & \multirow{2}{*}{0.8353} & \multirow{2}{*}{0.8236} & \multirow{2}{*}{0.8301} \\
(60 images) &  &  &  &  &  &  &  &  &  &  &  &  & \\\hline
  \end{tabular}
  }
  \label{tab:HDRTMQI}
\end{table*}
\begin{table*}[!t]
  \centering
  \caption{Experimental results for Simulation 1 (Statistical Naturalness)
  ``MEF,'' and ``CE'' indicate
  multi-exposure fusion and contrast enhancement, respectively.}
  {\footnotesize
  \begin{tabular}{l|l|lll|lll|ll|ll|ll} \hline \hline
    \multirow{3}{*}{Methods} & \multirow{3}{10mm}{Input image} & \multicolumn{3}{c|}{MEF} &
    \multicolumn{3}{c|}{CE} & \multicolumn{6}{c}{Proposed}\\\cline{3-14}
    & & \multicolumn{1}{c}{\cite{mertens2009exposure}} &
    \multicolumn{1}{c}{\cite{sakai2015hybrid}} & \multicolumn{1}{c|}{\cite{nejati2017fast}} &
    \multicolumn{1}{c}{HE} & \multicolumn{1}{c}{\cite{zuiderveld1994contrast}} &
    \multicolumn{1}{c|}{\cite{wu2017contrast}} &
    \multicolumn{2}{c|}{\cite{mertens2009exposure}} &
    \multicolumn{2}{c|}{\cite{sakai2015hybrid}} &
    \multicolumn{2}{c}{\cite{nejati2017fast}}\\
    &&&&&&&& \multicolumn{1}{c}{A} & \multicolumn{1}{c|}{B} &
    \multicolumn{1}{c}{A} & \multicolumn{1}{c|}{B} & \multicolumn{1}{c}{A} &
    \multicolumn{1}{c}{B}\\
    \hdashline
    AtriumNight & 0.1672 & 0.2185 & 0.2176 & 0.1644 & 0.3110 & 0.1398 & \textbf{0.4060} & 0.2411 & 0.2530 & 0.2398 & 0.2518 & 0.1829 & 0.1931 \\
    MtTamWest & 0.1972 & 0.2326 & 0.2328 & 0.2531 & 0.2231 & \textbf{0.7518} & 0.4140 & 0.3027 & 0.3781 & 0.2906 & 0.3612 & 0.2931 & 0.3681 \\
    SpheronNapa & 0.0116 & 0.0106 & 0.0105 & 0.0149 & 0.0418 & \textbf{0.1694} & 0.0720 & 0.0367 & 0.0430 & 0.0345 & 0.0403 & 0.0315 & 0.0368 \\
    Memorial & 0.2094 & 0.2113 & 0.2122 & 0.1945 & 0.2544 & 0.0444 & 0.2890 & 0.2311 & 0.2609 & 0.2367 & \textbf{0.2684} & 0.1935 & 0.2209 \\
    Rend 11 & 0.1637 & 0.2425 & 0.2365 & 0.2054 & 0.4703 & 0.5784 & \textbf{0.7145} & 0.2555 & 0.3645 & 0.2507 & 0.3576 & 0.2129 & 0.3197 \\\hdashline
    Average & \multirow{2}{*}{0.1498} & \multirow{2}{*}{0.1831} & \multirow{2}{*}{0.1819} & \multirow{2}{*}{0.1665} & \multirow{2}{*}{0.2601} & \multirow{2}{*}{0.3368} & \multirow{2}{*}{\textbf{0.3791}} & \multirow{2}{*}{0.2134} & \multirow{2}{*}{0.2599} & \multirow{2}{*}{0.2105} & \multirow{2}{*}{0.2558} & \multirow{2}{*}{0.1828} & \multirow{2}{*}{0.2277} \\
    (5 images) &  &  &  &  &  &  &  &  &  &  &  &  &  \\\hdashline
    Average & \multirow{2}{*}{0.2078} & \multirow{2}{*}{0.2000} & \multirow{2}{*}{0.2002} & \multirow{2}{*}{0.1903} & \multirow{2}{*}{0.3283} & \multirow{2}{*}{0.2683} & \multirow{2}{*}{\textbf{0.4496}} & \multirow{2}{*}{0.2543} & \multirow{2}{*}{0.2839} & \multirow{2}{*}{0.2528} & \multirow{2}{*}{0.2826} & \multirow{2}{*}{0.2278} & \multirow{2}{*}{0.2575} \\
    (60 images) &  &  &  &  &  &  &  &  &  &  &  &  & \\\hline
  \end{tabular}
  }
  \label{tab:HDRNaturalness}
\end{table*}
\begin{table*}[!t]
  \centering
  \caption{Experimental results for Simulation 1 (CIEDE2000)
  ``MEF,'' and ``CE'' indicate
  multi-exposure fusion and contrast enhancement, respectively.}
  {\small
  \begin{tabular}{l|r|rrr|rrr|rr|rr|rr} \hline \hline
    \multirow{3}{*}{Methods} & \multirow{3}{10mm}{Input image} & \multicolumn{3}{c|}{MEF} &
    \multicolumn{3}{c|}{CE} & \multicolumn{6}{c}{Proposed}\\\cline{3-14}
    & & \multicolumn{1}{c}{\cite{mertens2009exposure}} &
    \multicolumn{1}{c}{\cite{sakai2015hybrid}} & \multicolumn{1}{c|}{\cite{nejati2017fast}} &
    \multicolumn{1}{c}{HE} & \multicolumn{1}{c}{\cite{zuiderveld1994contrast}} &
    \multicolumn{1}{c|}{\cite{wu2017contrast}} &
    \multicolumn{2}{c|}{\cite{mertens2009exposure}} &
    \multicolumn{2}{c|}{\cite{sakai2015hybrid}} &
    \multicolumn{2}{c}{\cite{nejati2017fast}}\\
    &&&&&&&& \multicolumn{1}{c}{A} & \multicolumn{1}{c|}{B} &
    \multicolumn{1}{c}{A} & \multicolumn{1}{c|}{B} & \multicolumn{1}{c}{A} &
    \multicolumn{1}{c}{B}\\
    \hdashline
AtriumNight & 0.000 & 2.872 & 2.816 & 1.628 & 8.769 & 7.536 & 10.127 & 2.231 & 2.511 & 2.208 & 2.490 & \textbf{1.176} & 1.357 \\
MtTamWest & 0.000 & 3.881 & 3.864 & 2.715 & 5.875 & 4.994 & 5.869 & 1.891 & 3.832 & 1.879 & 3.826 & \textbf{1.335} & 2.806 \\
SpheronNapa & 0.000 & 4.565 & 4.561 & 2.821 & 4.204 & 8.724 & 5.024 & 2.346 & 2.627 & 2.334 & 2.617 & \textbf{1.472} & 1.794 \\
Memorial & 0.000 & 2.984 & 2.932 & 3.544 & 6.795 & 9.617 & 9.105 & 1.762 & 2.690 & \textbf{1.742} & 2.682 & 2.443 & 3.213 \\
Rend 11 & 0.000 & 3.447 & 3.403 & 2.947 & 7.418 & 7.343 & 8.766 & 2.892 & 5.582 & 2.862 & 5.560 & \textbf{2.212} & 4.827 \\\hdashline
Average & \multirow{2}{*}{0.000} & \multirow{2}{*}{3.550} & \multirow{2}{*}{3.515} & \multirow{2}{*}{2.731} & \multirow{2}{*}{6.612} & \multirow{2}{*}{7.643} & \multirow{2}{*}{7.778} & \multirow{2}{*}{2.224} & \multirow{2}{*}{3.448} & \multirow{2}{*}{2.205} & \multirow{2}{*}{3.435} & \multirow{2}{*}{\textbf{1.727}} & \multirow{2}{*}{2.800} \\
(5 images) &  &  &  &  &  &  &  &  &  &  &  &  &  \\\hdashline
Average & \multirow{2}{*}{0.000} & \multirow{2}{*}{3.353} & \multirow{2}{*}{3.326} & \multirow{2}{*}{2.433} & \multirow{2}{*}{7.527} & \multirow{2}{*}{7.397} & \multirow{2}{*}{8.785} & \multirow{2}{*}{2.417} & \multirow{2}{*}{3.434} & \multirow{2}{*}{2.400} & \multirow{2}{*}{3.424} & \multirow{2}{*}{\textbf{1.912}} & \multirow{2}{*}{2.839} \\
(60 images) &  &  &  &  &  &  &  &  &  &  &  &  & \\\hline
  \end{tabular}
  }
  \label{tab:HDRCIEDE}
\end{table*}
\begin{table*}[!t]
  \centering
  \caption{Experimental results for Simulation 2 (Statistical Naturalness)
  ``MEF,'' and ``CE'' indicate
  multi-exposure fusion and contrast enhancement, respectively.}
  {\footnotesize
  \begin{tabular}{l|l|lll|lll|ll|ll|ll} \hline \hline
    \multirow{3}{*}{Methods} & \multirow{3}{10mm}{Input image} & \multicolumn{3}{c|}{MEF} &
    \multicolumn{3}{c|}{CE} & \multicolumn{6}{c}{Proposed}\\\cline{3-14}
    & & \multicolumn{1}{c}{\cite{mertens2009exposure}} &
    \multicolumn{1}{c}{\cite{sakai2015hybrid}} & \multicolumn{1}{c|}{\cite{nejati2017fast}} &
    \multicolumn{1}{c}{HE} & \multicolumn{1}{c}{\cite{zuiderveld1994contrast}} &
    \multicolumn{1}{c|}{\cite{wu2017contrast}} &
    \multicolumn{2}{c|}{\cite{mertens2009exposure}} &
    \multicolumn{2}{c|}{\cite{sakai2015hybrid}} &
    \multicolumn{2}{c}{\cite{nejati2017fast}}\\
    &&&&&&&& \multicolumn{1}{c}{A} & \multicolumn{1}{c|}{B} &
    \multicolumn{1}{c}{A} & \multicolumn{1}{c|}{B} & \multicolumn{1}{c}{A} &
    \multicolumn{1}{c}{B}\\
    \hdashline
Arno & 0.0031 & 0.0264 & 0.0243 & 0.0360 & \textbf{0.2246} & 0.0448 & 0.1291 & 0.0095 & 0.0947 & 0.0092 & 0.0903 & 0.0072 & 0.1200 \\
Cave & 0.0006 & 0.0188 & 0.0174 & 0.0527 & \textbf{0.3231} & 0.0034 & 0.0070 & 0.0004 & 0.0009 & 0.0004 & 0.0011 & 0.0005 & 0.0001 \\
Chinese garden & 0.0772 & 0.1076 & 0.1141 & 0.1341 & \textbf{0.3460} & 0.0880 & 0.2298 & 0.1044 & 0.2267 & 0.1034 & 0.2552 & 0.0904 & 0.1739 \\
Corridor 1 & 0.0000 & 0.0000 & 0.0000 & 0.0000 & \textbf{0.3556} & 0.0000 & 0.0015 & 0.0000 & 0.2112 & 0.0000 & 0.2076 & 0.0000 & 0.2371 \\
Corridor 2 & 0.0000 & 0.0085 & 0.0077 & 0.0053 & \textbf{0.3031} & 0.0006 & 0.0473 & 0.0001 & 0.0854 & 0.0001 & 0.0817 & 0.0000 & 0.1066 \\
Estate rsa & 0.0049 & 0.0458 & 0.0411 & 0.0411 & 0.4502 & 0.1564 & \textbf{0.6606} & 0.0160 & 0.1910 & 0.0149 & 0.1850 & 0.0118 & 0.1641 \\
Kluki & 0.2843 & 0.3584 & 0.3388 & 0.2889 & 0.3526 & 0.4205 & \textbf{0.9720} & 0.3992 & 0.6323 & 0.3852 & 0.6151 & 0.3731 & 0.6129 \\
Laurenziana & 0.4360 & 0.3424 & 0.3261 & 0.3799 & 0.3967 & 0.6133 & \textbf{0.9213} & 0.5328 & 0.8753 & 0.5232 & 0.8799 & 0.4939 & 0.8344 \\
Lobby & 0.0006 & 0.0037 & 0.0032 & 0.0043 & 0.4276 & 0.0031 & 0.0206 & 0.0008 & 0.4635 & 0.0008 & \textbf{0.4733} & 0.0008 & 0.4448 \\
Mountains & 0.2867 & 0.0622 & 0.0563 & 0.0692 & 0.4029 & 0.6072 & \textbf{0.8669} & 0.2741 & 0.1514 & 0.2669 & 0.1483 & 0.3588 & 0.1774 \\
Ostrow tumski & 0.0055 & 0.0199 & 0.0176 & 0.0489 & 0.1545 & 0.0636 & 0.1955 & 0.0119 & 0.3626 & 0.0115 & 0.3478 & 0.0117 & \textbf{0.4887} \\
Window & 0.0020 & 0.0068 & 0.0065 & 0.0070 & 0.2777 & 0.0133 & 0.0397 & 0.0043 & 0.3515 & 0.0042 & 0.3401 & 0.0036 & \textbf{0.4653}\\\hline
  \end{tabular}
  }
  \label{tab:CameraNaturalness}
\end{table*}
\begin{table*}[!t]
  \centering
  \caption{Experimental results for Simulation 2 (CIEDE2000)
  ``MEF,'' and ``CE'' indicate
  multi-exposure fusion and contrast enhancement, respectively.}
  {\small
  \begin{tabular}{l|r|rrr|rrr|rr|rr|rr} \hline \hline
    \multirow{3}{*}{Methods} & \multirow{3}{10mm}{Input image} & \multicolumn{3}{c|}{MEF} &
    \multicolumn{3}{c|}{CE} & \multicolumn{6}{c}{Proposed}\\\cline{3-14}
    & & \multicolumn{1}{c}{\cite{mertens2009exposure}} &
    \multicolumn{1}{c}{\cite{sakai2015hybrid}} & \multicolumn{1}{c|}{\cite{nejati2017fast}} &
    \multicolumn{1}{c}{HE} & \multicolumn{1}{c}{\cite{zuiderveld1994contrast}} &
    \multicolumn{1}{c|}{\cite{wu2017contrast}} &
    \multicolumn{2}{c|}{\cite{mertens2009exposure}} &
    \multicolumn{2}{c|}{\cite{sakai2015hybrid}} &
    \multicolumn{2}{c}{\cite{nejati2017fast}}\\
    &&&&&&&& \multicolumn{1}{c}{A} & \multicolumn{1}{c|}{B} &
    \multicolumn{1}{c}{A} & \multicolumn{1}{c|}{B} & \multicolumn{1}{c}{A} &
    \multicolumn{1}{c}{B}\\
    \hdashline
Arno & 0.000 & 8.621 & 8.601 & 10.319 & 12.433 & 8.593 & 12.896 & 3.317 & 12.391 & 3.293 & 12.365 & \textbf{2.289} & 13.228 \\
Cave & 0.000 & 15.858 & 15.826 & 19.969 & 31.178 & 6.045 & 9.757 & 1.353 & 31.862 & \textbf{1.290} & 31.881 & 1.297 & 32.508 \\
Chinese garden & 0.000 & 11.954 & 11.882 & 10.922 & 16.282 & 13.556 & 15.954 & 2.594 & 15.706 & 2.470 & 15.660 & \textbf{2.294} & 15.231 \\
Corridor 1 & 0.000 & 3.794 & 3.785 & 2.551 & 40.235 & 6.738 & 19.344 & 1.347 & 36.950 & 1.335 & 36.948 & \textbf{0.944} & 37.685 \\
Corridor 2 & 0.000 & 22.179 & 22.164 & 19.810 & 30.185 & 9.368 & 24.636 & 3.377 & 27.568 & 3.364 & 27.558 & \textbf{1.812} & 28.086 \\
Estate rsa & 0.000 & 11.064 & 11.025 & 8.969 & 17.134 & 14.656 & 21.380 & 3.916 & 15.092 & 3.877 & 15.071 & \textbf{2.999} & 13.963 \\
Kluki & 0.000 & 11.081 & 11.017 & 5.740 & 3.103 & 12.403 & 12.160 & 2.457 & 5.412 & 2.389 & 5.356 & \textbf{1.870} & 4.945 \\
Laurenziana & 0.000 & 10.809 & 10.789 & 7.449 & 6.372 & 9.849 & 11.054 & 2.097 & 7.696 & 2.032 & 7.667 & \textbf{1.711} & 7.269 \\
Lobby & 0.000 & 8.552 & 8.520 & 8.232 & 33.087 & 7.074 & 16.938 & 1.339 & 31.463 & 1.312 & 31.457 & \textbf{1.022} & 31.529 \\
Mountains & 0.000 & 6.066 & 6.069 & 6.325 & 13.475 & 6.246 & 9.603 & 1.248 & 4.308 & 1.239 & 4.308 & \textbf{0.852} & 4.131 \\
Ostrow tumski & 0.000 & 7.077 & 7.032 & 8.297 & 9.976 & 8.562 & 11.694 & 2.114 & 15.677 & 2.089 & 15.667 & \textbf{1.795} & 17.287 \\
Window evaluative & 0.000 & 5.077 & 5.057 & 4.537 & 22.795 & 6.531 & 8.342 & 2.246 & 21.415 & 2.230 & 21.422 & \textbf{1.477} & 21.859\\\hline
  \end{tabular}
  }
  \label{tab:CameraCIEDE}
\end{table*}
\section{Conclusion}
  Our proposed method produces pseudo multi-exposure images from a
  single image and the use of a local contrast enhancement method
  improves their quality.
  The proposed method is done by utilizing the relationship
  between the exposure values and pixel values.
  Approaches A and B used in the proposed method enables us to avoid color distortions
  and to strongly enhance the image contrast, respectively.
  Approach B is available even when the exposure value of an input image is unknown,
  while approach A is only available when the exposure value is known.
  Experimental results showed that the proposed method can effectively enhances images
  as well as conventional multi-exposure image fusion methods,
  without multi-exposure images.
  In addition, the proposed approach A
  outperforms typical contrast enhancement methods
  in terms of the color distortion.
  On the other hand, approach B allows us to strongly enhance
  the contrast of images without loss of details,
  unlike conventional contrast enhancement methods.

%\bibliographystyle{ieicetr}% bib style
%\bibliography{../../0_bibliography/ref_ieicetr}% your bib database

%\begin{thebibliography}{99}% more than 9 --> 99 / less than 10 --> 9
%\bibitem{}
%\end{thebibliography}

%\profile{}{}
\profile{Yuma Kinoshita}{
  received his B.Eng. and M.Eng. degrees from Tokyo Metropolitan University,
  Japan, in 2016 and 2018, respectively.
  From 2018, he has been a Ph.D. student
  at Tokyo Metropolitan University.
  He received IEEE ISPACS Best Paper Award in 2016.
  His research interests are in the area of image processing.
  He is a student member of IEEE and IEICE.
}
%\profile{Taichi Yoshida}{
%  
%}
\profile{Sayaka Shiota}{
  received the B.E., M.E. and Ph.D. degrees in intelligence and
  computer science, Engineering and engineering simulation from
  Nagoya Institute of Technology, Nagoya, Japan in 2007, 2009 and 2012,
  respectively.
  From February 2013 to March 2014, she had worked at the Institute
  professor.
  In April of 2014, she joined Tokyo Metropolitan University as an
  Assistant Professor.
  Her research interests include statistical speech recognition and
  speaker verification.
  She is a member of the Acoustical Society of Japan (ASJ),
  the IEICE, and the IEEE.
}
\profile{Hitoshi Kiya}{
  received his B.Eng. and M.Eng. degrees
  from Nagaoka University of Technology, Japan, in 1980
  and 1982, respectively, and his D.Eng. degree from Tokyo
  Metropolitan University in 1987. In 1982, he joined Tokyo
  Metropolitan University as an Assistant Professor, where he became
  a Full Professor in 2000. From 1995 to 1996, he attended
  the University of Sydney, Australia as a Visiting Fellow. He
  was/is the Chair of IEEE Signal Processing Society Japan Chapter,
  an Associate Editor for IEEE Trans. Image Processing,
  IEEE Trans. Signal Processing and IEEE Trans. Information
  Forensics and Security, respectively.
  He also serves/served as the President
  of IEICE Engineering Sciences Society (ESS), the Editor-in-Chief
  for IEICE ESS Publications, and the President-Elect of APSIPA,
  His
  research interests are in the area of signal and image processing
  including multirate signal processing, and security for multimedia.
  He received the IWAIT Best Paper Award in 2014 and
  2015, IEEE ISPACS Best Paper Award in 2016,
  the ITE Niwa-Takayanagi Best Paper Award in 2012, the
  Telecommunications Advancement Foundation Award in 2011,
  the IEICE ESS Contribution Award in 2010, and the IEICE Best
  Paper Award in 2008. He is a Fellow Member of IEEE, IEICE and
  ITE.
}
% without picture of author's face

\end{document}